\documentclass[table]{article}
\usepackage{iclr2025_conference,times}

\usepackage[utf8]{inputenc} 
\usepackage[T1]{fontenc}    
\usepackage{hyperref}       
\usepackage{url}            
\usepackage{booktabs}       
\usepackage{nicefrac}       
\usepackage{microtype}      
\usepackage{pifont}
\usepackage{adjustbox}
\usepackage{subcaption}
\usepackage{enumitem}

\usepackage{amsmath}
\usepackage{amssymb}
\usepackage{mathtools}
\usepackage{latexsym}
\usepackage{dsfont}
\usepackage{mathrsfs}
\usepackage{amsfonts}
\usepackage{bm}
\usepackage{xspace}
\usepackage{amsthm}
\usepackage{multirow}

\usepackage{import}
\usepackage{pifont}
\usepackage{textcomp}
\usepackage{color, colortbl}
\definecolor{greyC}{RGB}{180,180,180}
\definecolor{greyL}{RGB}{235,235,235}
\usepackage{footmisc}
\usepackage[noend]{algpseudocode}
\usepackage{bm}
\usepackage[flushleft]{threeparttable}
\usepackage[nospace]{cite}
\usepackage{makecell}
\usepackage{wrapfig}

\usepackage{algorithm}
\usepackage{algpseudocode}

\newcolumntype{P}[1]{>{\raggedright\arraybackslash}p{#1}}  

\usepackage{hyperref} 
\hypersetup{
    colorlinks=true,   
    linkcolor=red,     
    citecolor=cyan,    
    filecolor=magenta, 
    urlcolor=magenta   
}

\definecolor{champagne}{RGB}{247, 231, 206} 
\definecolor{green(pigment)}{rgb}{0.0, 0.65, 0.31}
\definecolor{darksalmon}{rgb}{0.91, 0.59, 0.48}

\definecolor{mygray}{gray}{.92}

\definecolor{baselinecolor}{rgb}{1, 1, 1}

\definecolor{ourmethodcolor}{rgb}{0.94, 0.97, 1}

\definecolor{champagne}{RGB}{247, 231, 206} 

\usepackage{etoc}
\etocdepthtag.toc{mtchapter}
\etocsettagdepth{mtchapter}{subsection}
\etocsettagdepth{mtappendix}{none}

\definecolor{mblue}{RGB}{0, 61, 124}
\definecolor{myellow}{RGB}{239, 124, 0}
\definecolor{mnavy}{RGB}{0,0,128}
\definecolor{minc}{RGB}{0,128,0}
\definecolor{mdec}{RGB}{255,0,0}
\definecolor{mhold}{RGB}{128,128,128}

\definecolor{darksalmon}{rgb}{0.91, 0.59, 0.48}
\definecolor{emerald}{rgb}{0.31, 0.78, 0.47}
\definecolor{green(pigment)}{rgb}{0.0, 0.65, 0.31}
\definecolor{amaranth}{rgb}{0.9, 0.17, 0.31}
\definecolor{iris}{rgb}{0.35, 0.31, 0.81}
\definecolor{uu}{rgb}{0.95, 0.51, 0.51}
\definecolor{spirodiscoball}{rgb}{0.06, 0.75, 0.99}

\usepackage{makecell}

\theoremstyle{plain}

\theoremstyle{definition}

\theoremstyle{remark}

\usepackage[most]{tcolorbox}
\usepackage{float}
\usepackage{xspace}
\tcbset{
aibox/.style={
width=\textwidth/2,
top=10pt,
colback=white,
colframe=black,
colbacktitle=black,
enhanced,
center,
attach boxed title to top left={yshift=-0.12in,xshift=0.15in},
boxed title style={boxrule=0pt,colframe=white},
}
}
\newtcolorbox{AIbox}[2][]{aibox,title=#2,#1}

\usepackage{hyperref}
\usepackage{url}

\title{The Lottery LLM Hypothesis, Rethinking What Abilities Should LLM Compression Preserve?}

\author{\hspace{-1mm}
Zhenheng Tang$^{1}$ \quad Xiang Liu$^{2}$ \quad 
Qian Wang$^{3}$ \quad Peijie Dong$^{2}$ \quad \\
\bf Bingsheng He$^{3}$ \quad Xiaowen Chu$^{2,\dagger}$ \quad Bo Li$^{1,\dagger}$
\\
    $^1$ CSE, The Hong Kong University of Science and Technology \\
    $^2$ DSA, The Hong Kong University of Science and Technology (Guangzhou) \\
    $^3$ National University of Singapore \\
}

\iclrfinalcopy 
\begin{document}

\maketitle
\begingroup\renewcommand\thefootnote{$^{\dagger}$}
\footnotetext{Corresponding author (xwchu@ust.hk and bli@ust.hk).}
\endgroup

\begin{abstract}
Motivated by reducing the computational and storage costs of LLMs, model compression and KV cache compression have attracted much attention from researchers. However, current methods predominantly emphasize maintaining the performance of compressed LLMs, as measured by perplexity or simple accuracy on tasks of common sense knowledge QA and basic arithmetic reasoning. In this blog, we present a brief review of recent advancements in LLMs related to retrieval-augmented generation, multi-step reasoning, external tools, and computational expressivity, all of which substantially enhance LLM performance. Then, we propose a lottery LLM hypothesis suggesting that for a given LLM and task, there exists a smaller lottery LLM capable of producing the same performance as the original LLM with the assistance of multi-step reasoning and external tools. Based on the review of current progress in LLMs, we discuss and summarize the essential capabilities that the lottery LLM and KV cache compression must possess, which are currently overlooked in existing methods.
\end{abstract}

\section{Current Efforts on Compressing LLMs and KV Cache}

LLMs have demonstrated remarkable proficiency in natural language processing, enabling sophisticated interactions and understanding of human language~\citep{openai2023gpt4}. To learn the tremendous knowledge in the training datasets, the current advanced LLMs like GPT4~\citep{openai2023gpt4} and Llama3~\citep{touvron2023llama} have enormous parameters like $7 \sim 750$ billion. Training such an LLM requires extensive computational resources, often measured in enormous GPU days using advanced NVIDIA GPUs~\citep{touvron2023llama}. This results in substantial electricity consumption, impacting both economic and energy costs~\citep{samsi2023words, tangDVFS}, and raising concerns regarding sustainable computing~\citep{wilkins2024hybrid}. Furthermore, providing inference services for LLMs necessitates numerous GPUs and incurs additional energy costs~\citep{samsi2023words, tangDVFS}, making it a significant challenge for widespread deployment~\citep{10.1145/3620666.3651329}.

\textbf{Compression methods.} To this end, both academic researchers and industrial engineers are trying to compress model parameters and reduce the model into a smaller one while keeping its performance unchanged. The typical compression algorithm includes the pruning~\citep{Sun2023ASA_wanda, Frantar2023SparseGPTML, dongpruner} and quantization~\citep{Yao2022ZeroQuantEA, Dettmers2022TheCF, dong2024stbllm} of LLM parameters, and KV cache compression~\citep{zhang2024h2o, xiao2024efficient}. However, most of the current methods that compress LLMs and KV cache only show guaranteed performance of the perplexity on some basic language tasks like Wikitext2~\citep{merity2016pointer} and PTB~\citep{marcus1993building}, common sense knowledge QA tasks~\citep{hendrycks2021measuring, talmor-etal-2019-commonsenseqa} and the basic arithmetic reasoning tasks~\citep{cobbe2021training} in small-scale evaluation but not in the real-world industrial scenarios.

\textbf{Missed aspects.} Some recent studies show that the LLMs may lose their advanced crucial abilities under the compressions like the long-context retrieval, long-context generation and long-document reasoning and so on~\citep{jaiswal2024compressing}. Also, the long-context understanding ability of LLMs is significantly reduced under the KV cache compression~\citep{yuan-etal-2024-kv}.

In the following sections, we examine recent advancements in retrieval-augmented generation, the utilization of external tools, and multi-step reasoning, all of which markedly enhance the performance of LLMs. Subsequently, we introduce the lottery LLM hypothesis, which posits that for a specific LLM and task, a smaller lottery LLM can achieve equivalent performance to the original LLM, aided by multi-step reasoning and external tools. Drawing from the review of current LLM advancements, we discuss and outline the critical capabilities that the lottery LLM and KV cache compression should encompass, which are currently neglected in existing methodologies.

\section{Tackling Redundant and Unreal Knowledge of LLMs with Knowledge Retrieval}

\textbf{Redundant Knowledge.} In contemporary applications, many individuals utilize LLMs as encyclopedic resources or to verify news and academic research, akin to an Internet search engine. Recent studies indicate that LLMs exhibit varying performance in knowledge retrieval, contingent upon the popularity of the information~\citep{PopQA}. Specifically, a small subset of real-world question-answer (QA) pairs constitutes the majority of interactions, while a limited number of QAs receive frequent attention, demonstrating a long-tail distribution in their popularity~\citep{PopQA}. LLMs tend to perform better on high-popularity QAs compared to those with lower popularity.

\textbf{Hallucinated Knowledge.} LLMs often generate unreal outputs rather than factual knowledge, which is a phenomenon known as hallucination~\citep{huang2023survey}. This issue has garnered significant attention from researchers~\citep{huang2023survey}. There is ongoing debate regarding the feasibility of completely eliminating hallucinations~\citep{farquhar2024detecting}. Some studies suggest that hallucinations are inevitable, as they are a byproduct of the model's reasoning and generalization abilities~\citep{banerjee2024llms, xu2024hallucination}.

\textbf{Retrieval Augmented Generation (RAG).} Large Language Models (LLMs) exhibit robust in-context learning capabilities, enabling them to respond to queries using prompts rather than relying solely on their internal knowledge encoded within model parameters. Consequently, external knowledge sources such as scholarly articles, web pages, books, and other documents can be integrated into prompts to facilitate the retrieval of additional factual information~\citep{yao2022react}, thereby mitigating the occurrence of hallucinations~\citep{yao2022react}. This approach raises significant research questions:

\emph{Is it necessary to store all knowledge within LLM parameters if RAG can accurately retrieve factual information from external knowledge bases? If not, which knowledge should be stored and which should not?}


Considering two extreme scenarios:
\begin{itemize}[leftmargin=*]
\item  Storing all knowledge in \textbf{model parameters}: If all knowledge is stored within model parameters, LLMs function as oracle machines, obviating the need for RAG. However, training such an LLM is nearly impossible because not all knowledge can be collected and never outdated~\citep{xu2024hallucination, banerjee2024llms}. Moreover, deploying such a large model is inefficient.
\item  Storing all knowledge in \textbf{external knowledge bases}: If all knowledge is stored externally, LLM parameters could potentially be reduced significantly, allowing for the retrieval of factual information during inference.
\end{itemize}

Nevertheless, LLMs require foundational common knowledge to perform tasks such as reasoning and accurate retrieval. This issue will be further explored in subsequent sections. Thus, compressing all knowledge into external knowledge bases is not feasible. Investigating the nature of learned knowledge and identifying which knowledge triggers the grokking phenomenon in LLMs remains an open research question~\citep{Grokking}.

\textbf{Trade-off between model size and knowledge base.} Some studies indicate that adaptive knowledge retrieval is a promising direction to enhance the performance of LLMs and may help to find an optimal trade-off between the knowledge base and model size~\citep{jeong2024adaptive}. The adaptive RAG~\citep{soudani2024fine, jeong2024adaptive} suggests that popular knowledge can be stored in the model parameters, while less popular knowledge can be stored in the external knowledge base.

The core idea of adaptive RAG appears to be related to a classic efficient data structure, \textbf{Huffman coding}~\citep{moffat2019huffman}. Specifically, the cost of knowledge retrieval can be viewed as the prompt length (since the retrieved knowledge will be inserted into the prompts). Storing knowledge in the model parameters results in a shorter prompt length because LLMs can directly respond to questions without needing to retrieve knowledge from the external knowledge base. Conversely, storing knowledge in the external knowledge base results in a longer prompt length, implying higher retrieval operations and longer context lengths, which incur greater computational and storage costs during inference~\citep{xiao2024efficient}. Therefore, the popularity of the knowledge can be seen as the appearance probability, as in Huffman coding. Storing popular knowledge in the model parameters is more efficient.

\textbf{Finetuning vs. retrieval.} Another related question is whether finetuning should be used to enhance the performance of LLMs in specific application domains such as legal, finance, and medical fields~\citep{hendrycks2021measuring, talmor-etal-2019-commonsenseqa}. Finetuning may lead to the forgetting problem and additional training overheads, sparking debate on whether finetuning should be employed to improve LLM performance or if reliance on RAG can achieve the same goal~\citep{jeong2024adaptive}. Recent studies demonstrate that RAG can significantly enhance LLM performance in specific domains such as legal~\citep{pipitone2024legalbench}, medical~\citep{jeong2024improving}, and finance~\citep{li-etal-2024-alphafin}.

\textbf{Beyond the RAG.} Document-based knowledge retrieval primarily assists LLMs in retrieving knowledge of triplets consisting of entity, relation, and object~\citep{chen-etal-2024-retrieval}. However, the capabilities and exceptional performance of LLMs extend beyond retrieving triplet knowledge. LLMs also exhibit remarkable abilities such as solving arithmetic problems, playing chess, and coding, which are not simple triplet knowledge retrieval tasks~\citep{chen-etal-2024-retrieval}. Ensuring the reasoning performance of smaller LLMs is crucial and cannot be easily addressed by document-based knowledge retrieval.

\section{External Tools}

Advanced Large Language Models (LLMs) demonstrate remarkable capabilities in function calling, which involves invoking external tools to address specific tasks. These external tools may include Internet search engines~\citep{ToolLLM}, arithmetic calculation functions~\citep{NEURIPS2023_d842425e}, system operations~\citep{LLM-as-OS, AIOS}, game interfaces, and more. These are formulated into programming function calls~\citep{Granite-Function} and conveyed to LLMs via prompts. Based on the function descriptions, LLMs determine which function to call to resolve the given problems~\citep{Granite-Function}.

\textbf{Arithmetic Function Calls.} To solve arithmetic problems, LLMs are trained on arithmetic datasets~\citep{cobbe2021training}. However, simple errors often occur during the arithmetic reasoning process, such as LLMs erroneously determining that 9.11 is greater than 9.9~\citep{choi2024automatic}. To mitigate this, some studies propose enabling LLMs to generate programs that include arithmetic operations and utilize an external Python interpreter to solve these problems~\citep{pmlr-v202-gao23f}. Additionally, some research suggests leveraging arithmetic function calls to solve arithmetic problems~\citep{he23solving}. Experimental results indicate that arithmetic function calling can significantly enhance the performance of LLMs on arithmetic tasks~\citep{pmlr-v202-gao23f, NEURIPS2023_e3936777}.

\textbf{Internet Search Engine.} To augment LLM knowledge with online and dynamically updated external information, the Internet search engine is employed as an external tool~\citep{yao2022react, FreshLLMs}. Experimental results demonstrate that interacting with an Internet search engine, such as a simple Wikipedia API, can significantly improve LLM performance on knowledge retrieval tasks~\citep{yao2022react}.

\textbf{LLM Operating System (OS).} By conceptualizing LLM calls as system calls akin to traditional operating systems, recent studies propose developing a new \textit{LLM-as-OS} framework~\citep{LLM-as-OS}, which allows LLMs to invoke external tools like applications in an OS. Recent studies also propose the AIOS framework~\citep{AIOS} to decouple LLM calls from system calls and implement various managers to enhance AIOS efficiency. The optimized agent framework from the OS perspective significantly improves both the efficiency and performance of LLM calls.

\textbf{Logic Solver.} There is ongoing debate regarding whether LLMs can perform logical reasoning akin to humans~\citep{GSM-Symbolic, CanLLMReason, valmeekam2022large, 10.1145/3627673.3679832, xu2023large, arkoudas2023gpt4cantreason}. Recent studies suggest that to enhance the reasoning capabilities of LLMs, external logic solvers can be utilized to solve logical reasoning problems~\citep{RecursiveReasoning}. In some frameworks, LLMs are tasked with transforming natural language sentences into logical forms, while logic solvers are responsible for solving the logical reasoning problems~\citep{han2022folio, pan-etal-2023-logic, wang2024symbolic}. Other frameworks propose allowing LLMs to summarize sentences into premises and conclusions, then aggregate this extracted information into another prompt to enable Logic inference~\citep{sun-etal-2024-determlr, wang2024symbolic, Xu2024FaithfulLR}.

\section{Computational Expressivity of LLMs}

\textbf{Basic Transformer Architecture.} Basic transformers, devoid of intermediate decoding steps, exhibit limited computational expressivity~\citep{10.1162/tacl_a_00562, chiang2023tighter}, aligning with the relatively small circuit complexity class $TC^0$~\citep{10.1162/tacl_a_00562}. These basic transformers fall short of Turing completeness, as they are incapable of solving problems that are complete for classes larger than $TC^0$, such as simulating automata, which is $NC^1$-complete.

\textbf{Decoding-based Transformers.} Decoding-based transformers generate output sequentially, word by word, rather than producing a single answer. This approach enhances their computational expressivity compared to basic transformers, with expressivity increasing in tandem with the length of the decoding steps~\citep{merrillexpressive}. This phenomenon elucidates why the Chain-of-Thought (CoT) reasoning process~\citep{CoT} augments the computational expressivity of LLMs~\citep{Reveal-CoT}. Some studies demonstrate that with linear steps, transformers equipped with projected-norm can theoretically simulate a Turing automaton~\citep{merrillexpressive}. Recent research indicates that autoregressive decoding, which facilitates the processing of arbitrarily long input strings, can simulate a universal Turing machine~\citep{Autogressive-Turing}.

\textbf{Decoding with External Memory.} Research suggests that external memory can enhance the computational expressivity of LLMs~\citep{deletang2023neural}, potentially endowing them with approximate Turing completeness~\citep{JMLRv2220-302}. Recent advancements have introduced the Stack-Attention mechanism to further augment the reasoning capabilities of LLMs~\citep{Stack-Attention}. With the integration of external memory and simple regular expression parsers, transformers can simulate the execution of a universal Turing machine, specifically $U_{15,2}$~\citep{Memory-Augmented-Turing}.

\section{Multi-step Reasoning}

The Chain-of-Thought (CoT) reasoning paradigm demonstrates that engaging in detailed, step-by-step reasoning can significantly enhance the performance of Large Language Models (LLMs) compared to single-step reasoning~\citep{CoT}. This improvement arises because single-step reasoning may overlook crucial intermediate steps that are instrumental in problem-solving~\citep{CoT}. The multi-step reasoning process, inspired by human cognitive processes, can substantially elevate the performance of LLMs~\citep{CoT}.

\textbf{Single LLM Call.} CoT exemplifies a single LLM call, utilizing the model once. Beyond explicit prompting to initiate detailed reasoning, recent studies propose enabling LLMs to execute advanced search algorithms during the decoding process, such as Monte-Carlo Tree Search (MCTS)~\citep{Decoding} or Q-star search~\citep{chakraborty2024transfer}. Additionally, some research suggests employing backtracking algorithms to allow LLMs to reconsider previous decisions, thereby enhancing final performance~\citep{fubreak}.

\textbf{Multiple LLM Calls.} Some approaches advocate for multiple LLM calls, which operate independently of each other, potentially yielding correct answers across these calls~\citep{brown2024large}. Beyond the single CoT call, CoT-SC proposes multiple CoT-based LLM calls, selecting the optimal answer to improve final outcomes~\citep{wangself}. However, these answers exhibit direct dependencies. To optimize scheduling and decomposition of the reasoning process, Tree-of-Thought (ToT) reasoning~\citep{ToT} and Graph-of-Thought (GoT) reasoning~\citep{GoT} have been introduced, structuring reasoning steps in tree-like or graph-like configurations. Some studies also suggest integrating knowledge graphs, enabling LLMs to reason within graph structures to enhance reasoning capabilities~\citep{luoreasoning, sunthink}. Structuring prompts into triplets using LLMs can further bolster reasoning abilities~\citep{jiang2023structgpt}. In the absence of a centralized controller, some research proposes simulating multiple agents with LLMs to collaboratively address problems~\citep{li2023camel, hong2024metagpt, liang2023encouraging, duimproving}.

\textbf{Planning and Scheduling.} The essence of multi-step reasoning lies in decomposing the original problem into multiple sub-problems and addressing them sequentially. This process involves planning and scheduling. To facilitate autonomous planning and scheduling, recent studies propose employing LLMs as meta-agents to orchestrate planning and scheduling, wherein the original problem is decomposed, and the meta-agent delegates sub-problems to other LLMs based on the schedule~\citep{hong2024metagpt, wu2024autogen, zhoulanguage, wangvoyager}. With the aid of external symbolic reasoning, LLMs can also engage in planning and scheduling to resolve problems~\citep{RecursiveReasoning}.

\section{Lottery LLM Hypothesis}

Consider an original language model $f_\theta$ parameterized by the $\theta \in \mathbb{R}^{k_{\theta}}$, capable of processing input of token length $n$, and an input problem $q \in \mathbb{R}^{m\times h}$ with token length $m < n$ and ground truth $\mu \in \mathbb{R}^{l\times h}$. The problem $q$ is a question consisting of a sequence of words. And the $\mu$ is also a sequence of words representing the answer to the question $q$. $h$ is the dimension of the word embedding. The performance of the model is evaluated using a performance measure $P(\cdot)$, expressed as $P(f_\theta(q), \mu)$ which map its inputs as a scalar value. We hypothesize the existence of a smaller language model $g_\phi$ with parameters $\phi \in \mathbb{R}^{k_{\phi}}$ ($k_{\phi} < k_{\theta}$) and the same input length $n$, which can solve the problem $q$ with performance comparable to $f_\theta$, such that:

\begin{equation}
P(f_\theta(q), \mu) \leq P( \mathcal{A}_{g_\phi, \mathcal{D}, \mathcal{R}, \mathcal{C}, \mathcal{M}}(q), \mu),
\end{equation}

\begin{wrapfigure}{R}{0.5\textwidth}
\begin{minipage}{0.5\textwidth}
\begin{figure}[H]
    \centering
    \includegraphics[width=1.0\textwidth]{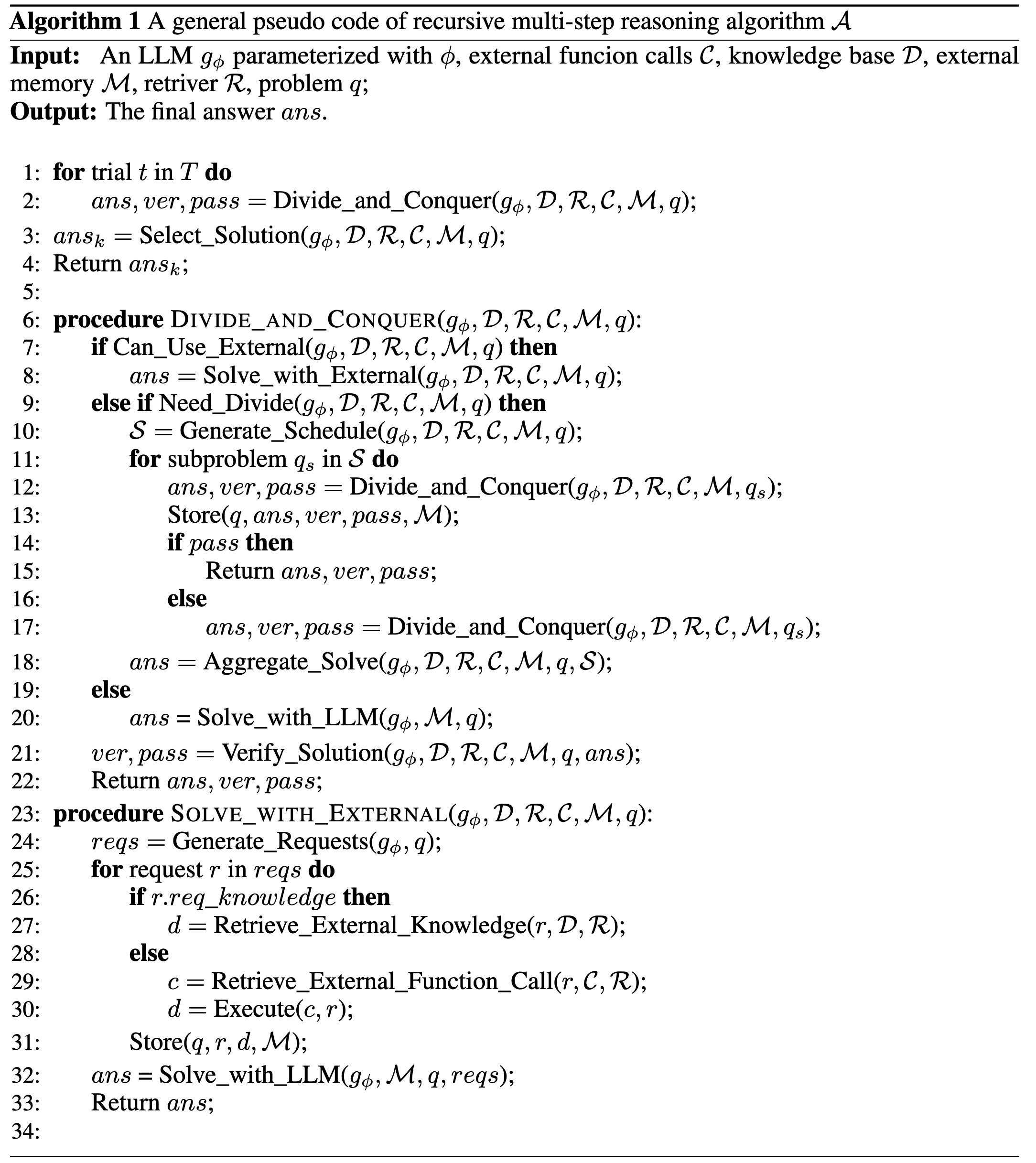}
    \caption{A general pseudo code of the reasoning algorithm $\mathcal{A}$.}
    \label{fig:algo}
\end{figure}
\end{minipage}
\end{wrapfigure}
where $\mathcal{A}$ represents a reasoning algorithm that may involve one or multiple invocations of $g_\phi$ with various inputs, including the original problem $q$, documents $d \in \mathcal{D}$ retrieved from the external knowledge base $\mathcal{D}$, or function calls $c \in \mathcal{C}$ retrieved from external tools $\mathcal{C}$ using the retriever $\mathcal{R}$. Each document $d \in \mathbb{R}^{n_d\times h}$ is a vector of words. While the function calls $c: \mathbb{R}^{n_c^i\times h} \to \mathbb{R}^{n_c^o\times h}$ is a provided function. The knowledge base $\mathcal{D}$ is a vector database storing vector-documents as key-value pairs, and $\mathcal{M}$ denotes the external memory that stores intermediate results. All $\mathcal{D}$, $\mathcal{C}$, and $\mathcal{M}$ are sets. And items in $\mathcal{D}$ and $\mathcal{C}$ are key-value pairs depends on the specific tasks, like vector database~\citep{pan2024survey}. The retriever $\mathcal{R}$ is a function that retrieves the required documents or function calls from the $\mathcal{D}$ or $\mathcal{C}$ based on the request. And its specific implementation can be various~\citep{gao2023retrieval}.

\begin{wrapfigure}{R}{0.5\textwidth}
\begin{minipage}{0.5\textwidth}
\begin{figure}[H]
    \centering
    \includegraphics[width=1.0\textwidth]{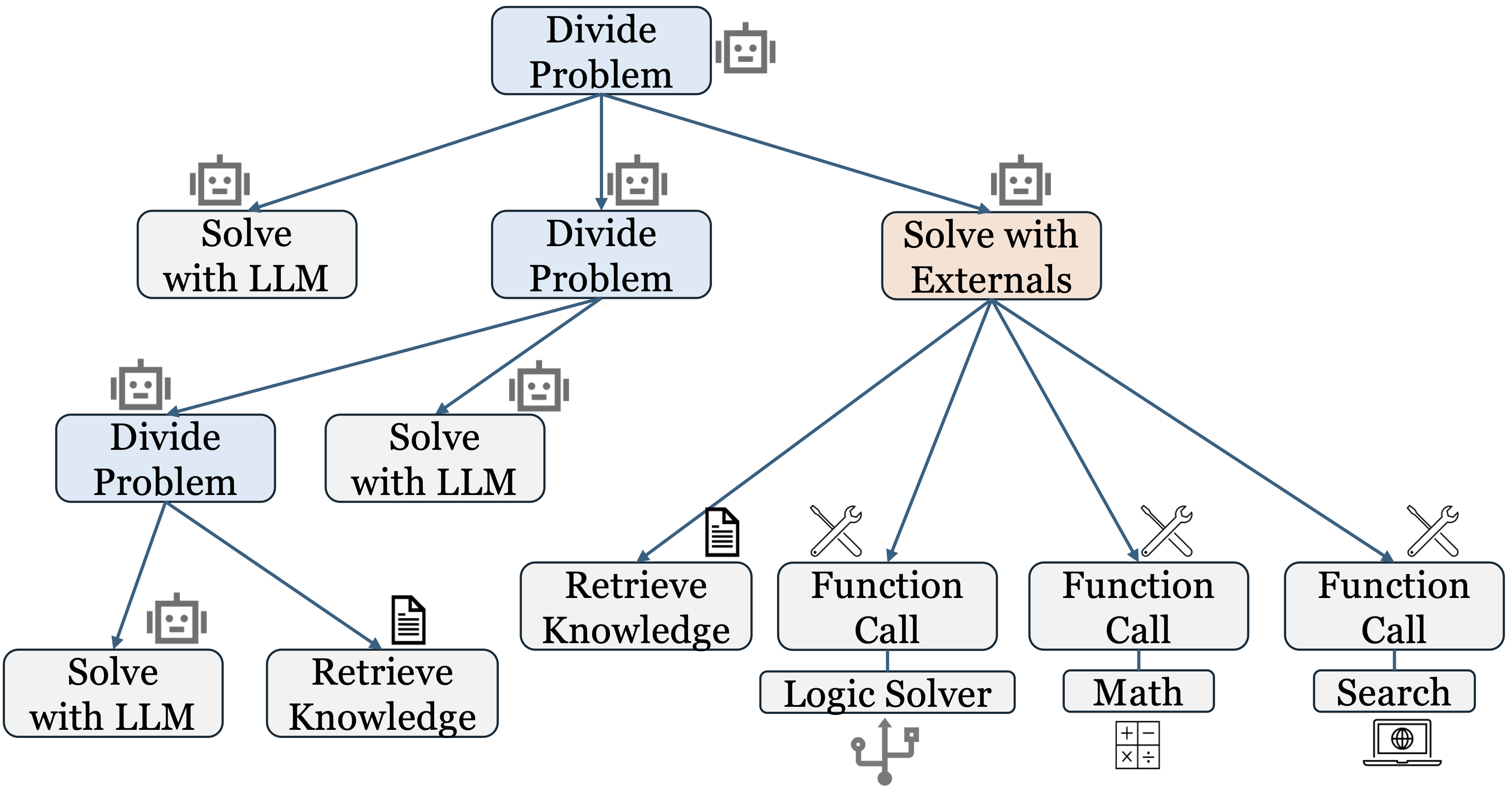}
    \caption{The problem solving process of the multi-step reasoning with external tools (the interaction with the external memory and the verification are not shown in the figure).}
    \label{fig:graph-structure}
\end{figure}
\end{minipage}
\end{wrapfigure}

The reasoning algorithm $\mathcal{A}$ is described as Algorithm 1 in Figure~\ref{fig:algo} and Figure~\ref{fig:graph-structure}, employing a divide-and-conquer strategy to solve the original problem $q$. This dynamic divide-and-conquer methodology is versatile and applicable to numerous contemporary reasoning algorithms.

\textbf{Recursive and Dynamic Scheduling.} Algorithm 1 can encompass tree-based reasoning methods such as Tree-of-Thought (ToT)~\citep{zhoulanguage, ToT}, due to its recursive design that facilitates tree search and allows the branch-or-solve mechanism to be dynamically determined by LLMs. Additionally, Algorithm 1 is applicable to graph-based reasoning methods like Graph-of-Thought (GoT)~\citep{GoT, luoreasoning, sunthink}, as the interaction between different LLMs and the external memory $\mathcal{M}$ can be conceptualized as a combination in GoT, where outputs from various nodes are integrated to construct the graph structure.

\textbf{External Knowledge and Tools.} During each phase of problem-solving, Algorithm 1 initially assesses whether the problem can be directly addressed using the external knowledge base $\mathcal{D}$ or external tools $\mathcal{C}$. If so, Algorithm 1 utilizes $g_\phi$ to evaluate the problem $q$ and ascertain the necessary knowledge or tools required for its resolution. Subsequently, based on the generated requests, the retriever $\mathcal{R}$ searches for external knowledge $d \in \mathcal{D}$ or tool $c \in \mathcal{C}$ to provide the requisite results. These supplementary results are then integrated with the problem $q$ for resolution by the model $g_\phi$. This framework facilitates the application of Retrieval Augmented Generation (RAG) and external tools, such as arithmetic calculation functions, Internet search engines, and logic solvers, to effectively address the problem $q$.

\begin{wrapfigure}{R}{0.3\textwidth}
\begin{minipage}{0.3\textwidth}
\begin{figure}[H]
    \centering
    \includegraphics[width=1.0\textwidth]{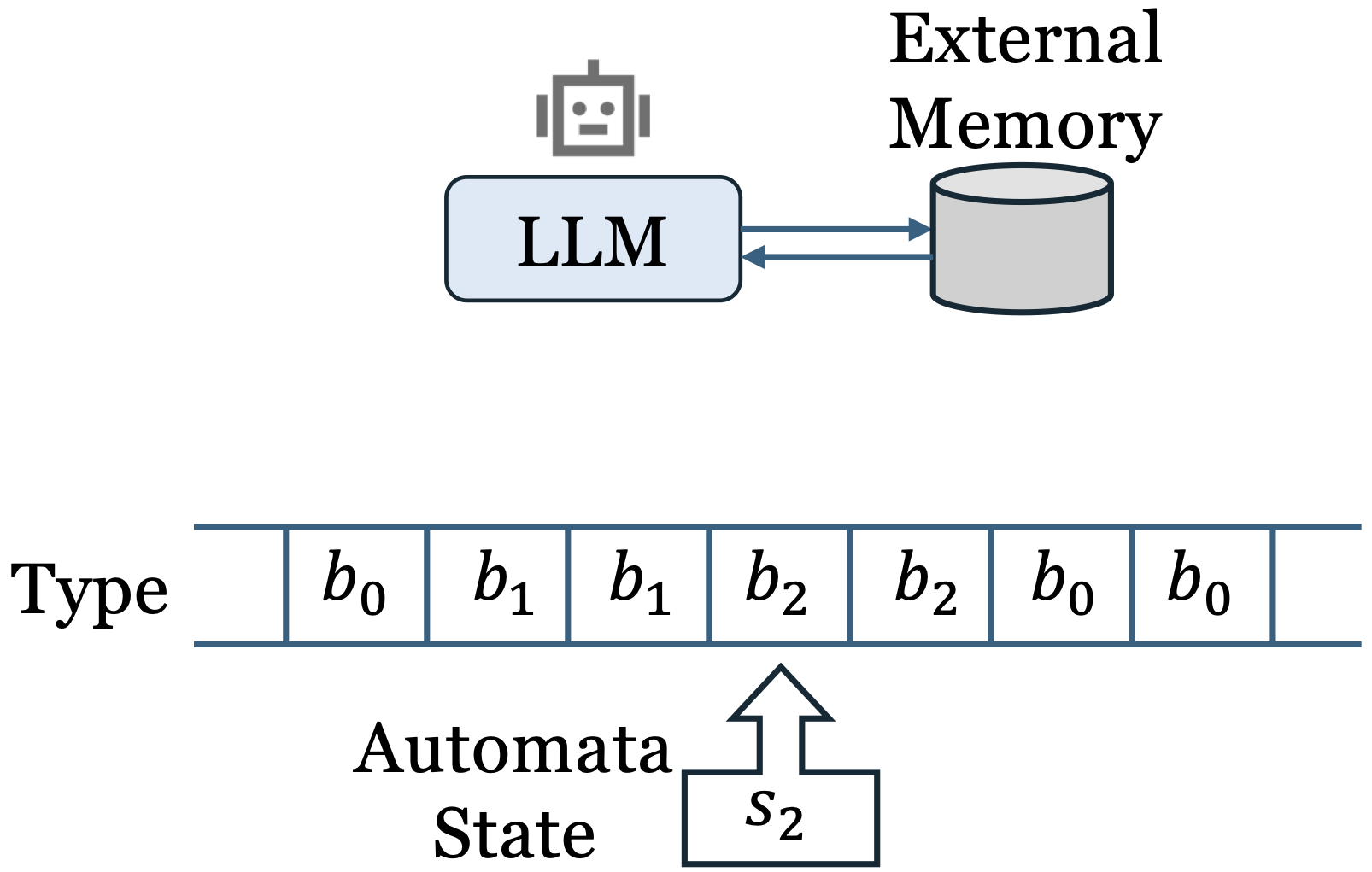}
    \caption{Simulating the Turing machine with LLMs and the external memory~\citep{Memory-Augmented-Turing}.}
    \label{fig:automate}
\end{figure}
\end{minipage}
\end{wrapfigure}

\textbf{External Memory.} The external memory $\mathcal{M}$ functions as a repository for storing intermediate results throughout the reasoning process. When tackling various sub-problems, intermediate results can be stored in the external memory for reuse in subsequent steps. By interacting with the external memory, Algorithm 1 can emulate reasoning methods that utilize working memory~\citep{wang2024symbolic}. The structure of the \textbf{Divide and Conquer} function in Algorithm 1 is not constrained. Through careful design and programming, the recursive mechanism can execute fundamental operations such as MOV, COPY, JUMP, and WRITE and READ from the external memory, thereby simulating a Turing machine~\citep{Memory-Augmented-Turing}, as depicted in Figure~\ref{fig:automate}.

Most of previous model compression~\citep{Sun2023ASA_wanda, Frantar2023SparseGPTML}~\citep{Yao2022ZeroQuantEA, Dettmers2022TheCF} and KV cache compression methods~\citep{zhang2024h2o, xiao2024efficient} only focus on the guaranteeing the model performance on the perplexity metric~\citep{merity2016pointer} or some downstream tasks like the common sense knowledge~\citep{hendrycks2021measuring, talmor-etal-2019-commonsenseqa} and the basic arithmetic problems~\citep{cobbe2021training}. From the above analysis and the procedures of the Algorithm 1, we can see that there are some other crucial abilities that the lottery LLM and other compression methods must take for considering. We summarize the crucial abilities that the lottery LLM should have as follows.

\textbf{Ability 1: Retrieval from prompts.} Obviously, the useful information in the prompts that related to address the problem $q$ is crucial for the lottery LLM. After collecting the required external results into the prompt, the LLM $g_\phi$ needs to be able to retrieve the required information from the prompt and avoid the interruption of some irrelevant information. This is related to the retrieval ability of the LLM and its measurement test is like the well-known needle-in-the-haystack(NIAH) test~\citep{needle}. We show that there is a simple and interesting method to endow the LLM with advanced retrieval ability with preprocessing prompts, by applying a the embedding to retrieve the related information about the question in problem $q$ and combine them with the question to prompt the LLM $g_\phi$ rather let the LLM $g_\phi$ to process the original long context information of problem $q$.

\begin{figure*}[h]
    \centering
    \includegraphics[width=1.0\textwidth]{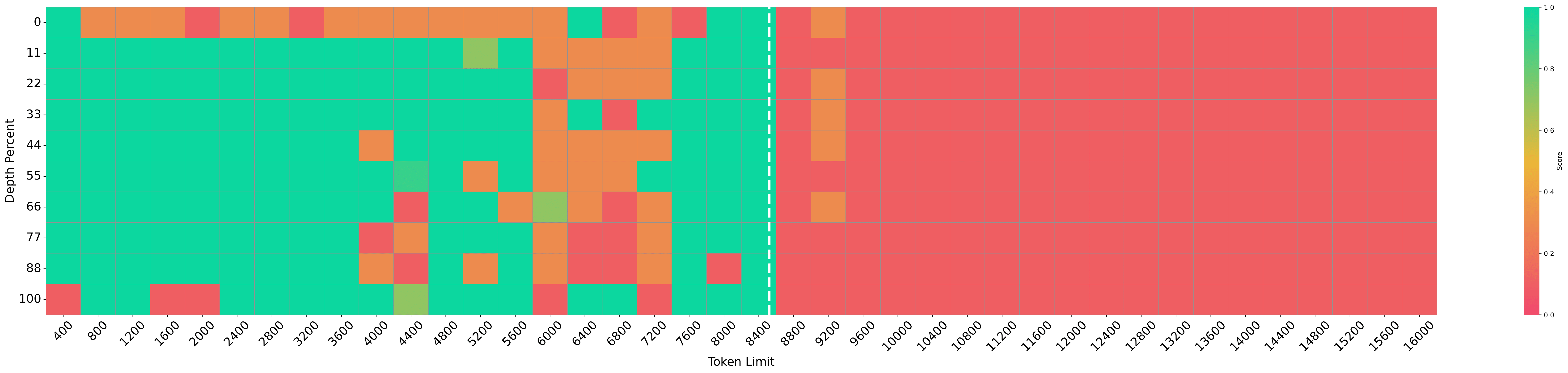}
    \caption{Vanilla NIAH results of LLaMA3-8B-Instruct.}
\end{figure*}

\begin{figure*}[h]
    \centering
    \includegraphics[width=1.0\textwidth]{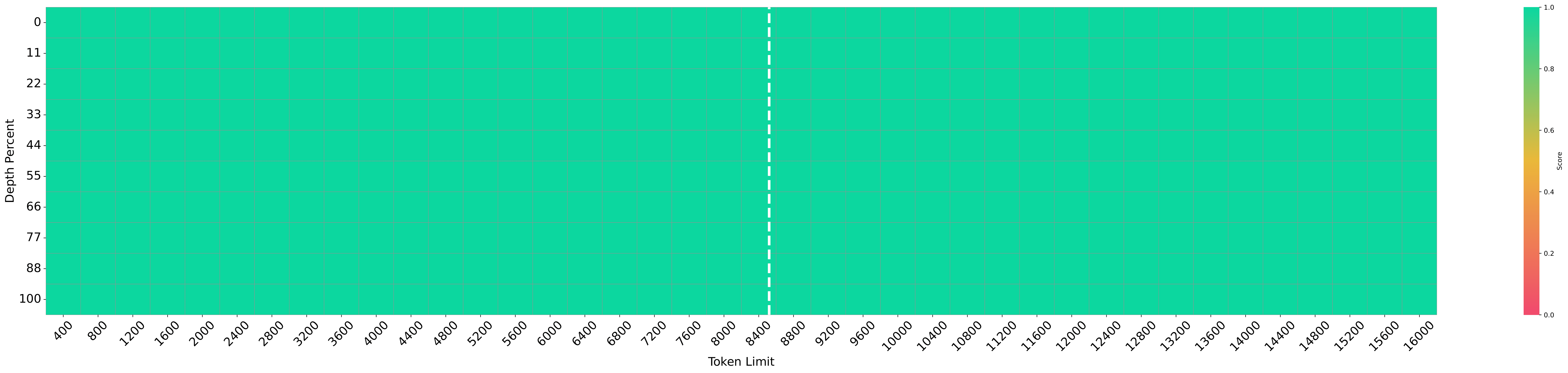}
    \caption{NIAH results of LLaMA3-8B-Instruct with preprocessing prompts.}
\end{figure*}

The figures illustrate that preprocessing prompts markedly enhances the performance of LLMs on the NIAH test. Importantly, even when the input length surpasses the model's context size (8K tokens for LLaMA3-8B-Instruct), there is no observed degradation in performance. This indicates the potential of utilizing preprocessed prompts to augment the retrieval capabilities of LLMs.

\textbf{Ability 2: Identification of Required External Resources.} To effectively determine which external resources to utilize, such as knowledge databases or external tools, the LLM $g_\phi$ must possess the capability to comprehend and correlate the problem $q$ and its associated sub-problems with the relevant resources. Consequently, $g_\phi$ should have foundational knowledge of the problem $q$ and the external resources. Additionally, it must exhibit a strong ability to associate queries with the available resources. When external tools are adeptly employed, the performance of smaller LLMs can be significantly enhanced. The subsequent table presents the results of arithmetic problem-solving using various LLMs and methodologies. The PAL~\citep{pmlr-v202-gao23f} approach, which employs external arithmetic calculation functions, demonstrates a substantial improvement in the performance of smaller LLMs.

\begin{table*}[h]
\centering
\resizebox{0.7\linewidth}{!}{ 
\begin{tabular}{|l|c|c|c|c|c|}
\hline
        & GSM8K & SVAMP & ASDIV &  ADDSUB & MULTIARITH \\
\hline
DIRECT Codex & 19.7  &69.9  & 74.0  &  90.9   & 44.0       \\
CoT UL2-20B  & 4.1   &12.6  & 16.9  & 18.2   & 10.7       \\
CoT LaMDA-137B & 17.1  & 39.9  & 49.0  &  52.9   & 51.8       \\
CoT Codex    & 65.6  & 74.8  & 76.9  &  86.0   & 95.9       \\
CoT PaLM-540B & 56.9  & 79.0  & 73.9  &  91.9   & 94.7       \\
CoT Minerva 540B & 58.8  & -     & -     & -      & -          \\
PAL~\citep{pmlr-v202-gao23f}            & \textbf{72.0}  & \textbf{79.4}  & \textbf{79.6}  & \textbf{92.5}   & \textbf{99.2}       \\
\hline
\end{tabular}
}
\caption{Arithmetic problem-solving results using various LLMs and methodologies.}
\end{table*}

Besides, with provided the external documents, following results~\citep{asai2024selfrag} show that the small LLM (Llama-3-Ins8B) show the superb performance in many QA tasks~\citep{mallen2023not, kwiatkowski2019natural, stelmakh2022asqa} than the large LLMs (Llama-3-Ins70B and ChatGPT-4oMINI).

\begin{table*}[h]
\centering
\resizebox{0.7\linewidth}{!}{ 
\begin{tabular}{|l|l|c|c|c|c|}
\hline
Method & LLM & PopQA (acc) & NQ (acc) & ASQA (str-em) & ASQA (hit) \\
\hline
CoT without RAG & Llama-3-Ins8B & 24.8 & 44.0 & 28.8 & 7.8 \\
CoT without RAG  & Llama-3-Ins70B & 31.6 & 54.4 & 36.4 & 11.2 \\
CoT without RAG & ChatGPT-4oMINI & 32.4 & 53.2 & 32.4 & 8.0 \\
With RAG & Llama-3-Ins8B & \textbf{59.8} & \textbf{54.0} & \textbf{38.8} & \textbf{14.0} \\
\hline
\end{tabular}
}
\caption{QA task performance with and without RAG.}
\end{table*}

\textbf{Ability 3: Planning and Scheduling.} To effectively decompose the problem $q$ into multiple sub-problems and address them sequentially, the LLM $g_\phi$ must possess robust planning and scheduling capabilities. This competency is essential for the lottery LLM to tackle complex problems efficiently. Consequently, the LLM $g_\phi$ should have a comprehensive understanding of both the primary problem $q$ and its constituent sub-problems. However, the intricate details of solving these sub-problems may not be necessary for the LLM $g_\phi$, as external resources can be leveraged to resolve them. Moreover, proficient scheduling is crucial for the lottery LLM to enhance reasoning efficiency.

The table below illustrates the performance of LLMs using simple inference compared to those employing a strategy of decomposing the problem into sub-problems and utilizing external logic solvers, such as Logic-LM~\citep{pan-etal-2023-logic}. The used five datasets are commonly used in the logical reasoning tasks. Notably, we emphasize the results~\citep{pan-etal-2023-logic} (simple inference/with Logic-LM) of GPT-3.5, which, despite being less advanced than GPT-4, demonstrates comparable performance to GPT-4 (GPT-3.5 with Logic-LM compared with GPT-4 with simple inference). Thus, with advanced reasoning algorithms, the weaker LLMs can outperform the stronger LLMs in advanced tasks.

\begin{table*}[h]
\centering
\resizebox{0.7\linewidth}{!}{ 
\begin{tabular}{|l|c|c|c|}
\hline
Dataset            & ChatGPT (gpt-3.5-turbo) & GPT-3.5 (text-davinci-003) & GPT-4 (gpt-4) \\
\hline
PrOntoQA           & 47.40 / 61.00           & 51.80 / \textbf{85.00}          & 77.40 / 83.20 \\
ProofWriter        & 35.50 / 58.33           & 36.16 / \textbf{71.45}          & 52.67 / 79.66 \\
FOLIO              & 45.09 / \textbf{62.74}           & 54.60 / 61.27          & 69.11 / 78.92 \\
LogicalDeduction   & 40.00 / \textbf{65.67}           & 41.33 / 62.00              & 71.33 / 87.63 \\
AR-LSAT            & 20.34 / 26.41           & 22.51 / \textbf{25.54}              & 33.33 / 43.04 \\
\hline
\end{tabular}
}
\caption{Performance of LLMs using simple inference and Logic-LM~\citep{pan-etal-2023-logic}.}
\end{table*}

\textbf{Ability 4: Precise Approximation of Fundamental Operations.} As discussed in the section on the computational expressivity of LLMs, achieving (approximate) Turing completeness necessitates that the LLM $g_\phi$ precisely approximates fundamental operations such as MOV, COPY, JUMP, and WRITE and READ from external memory~\citep{Autogressive-Turing, Memory-Augmented-Turing}. Although these operations may not be directly employed in problem-solving, they are essential for the lottery LLM to function as a potential meta-agent~\citep{hong2024metagpt}.

\textbf{Ability 5: Long-Context Reasoning.} In single-step reasoning, an extended context length allows the LLM $g_\phi$ to access and utilize more information for problem-solving. In multi-step reasoning, the prompt serves as a form of working memory for the meta-agent, or planner (controller). Each result from solved sub-problems should be incorporated into the prompt for subsequent steps. As problem complexity increases, so does the depth of the sub-problem tree. Therefore, the LLM $g_\phi$ must possess the ability for extended contextual reasoning to support deep tree reasoning~\citep{merrillexpressive, Reveal-CoT}.

\section{Conclusion}

This blog aims to elucidate the potential of the lottery LLM and to summarize the essential capabilities that the lottery LLM should possess, which are currently lacking in existing methods of LLM and KV cache compression. The discussion on redundant knowledge within LLMs also highlights the trade-off between knowledge storage and reasoning capabilities. With the development of the lottery LLM, alongside external tools, knowledge bases, and a robust algorithm $\mathcal{A}$, there is potential for the lottery LLM to function as a meta-agent akin to human cognition. Its external memory could serve as long-term memory, the prompt as short-term memory, and the LLM inference process $g_\phi$ as the fundamental cognitive process. External tools and knowledge bases can be considered as supplementary tools commonly used in daily life. Deploying the lottery LLM could significantly reduce energy and resource consumption in large-scale LLM-driven applications. Future research on LLM compression, KV cache compression, and other efficient LLM methodologies should address both efficiency and the essential capabilities of LLMs.

\section*{ACKNOWLEDGMENT}



This work was partially supported by National Natural Science Foundation of China under Grant No. 62272122, the Guangzhou Municipal Joint Funding Project with Universities and Enterprises under Grant No. 2024A03J0616, Guangzhou Municipality Big Data Intelligence Key Lab (2023A03J0012), and Hong Kong CRF grants under Grant No. C7004-22G and C6015-23G, contract R6021-20, and RGC GRF grants under the contracts 16200221, 16207922 and 16207423, the MOE Academic Research Fund (AcRF) Tier 1 Grant in Singapore (Grant No. T1 251RES2315).


\bibliography{iclr2025_conference}
\bibliographystyle{iclr2025_conference}


\end{document}